\def\year{2021}\relax
\pdfoutput=1
\documentclass[letterpaper]{article} % DO NOT CHANGE THIS
\usepackage{aaai21}
\usepackage[switch]{lineno}
\usepackage{amsmath}% DO NOT CHANGE THIS
\usepackage{times}  % DO NOT CHANGE THIS
\usepackage{helvet} % DO NOT CHANGE THIS
\usepackage{courier}  % DO NOT CHANGE THIS
\usepackage[hyphens]{url}  % DO NOT CHANGE THIS
\usepackage{graphicx} % DO NOT CHANGE THIS
\usepackage{natbib}  % DO NOT CHANGE THIS AND DO NOT ADD ANY OPTIONS TO IT
\usepackage{caption} % DO NOT CHANGE THIS AND DO NOT ADD ANY OPTIONS TO IT
\title{A Cross-lingual Natural Language Processing Framework for Infodemic Management}
\author {
        Ridam Pal,\textsuperscript{\rm 1 \#}
        Rohan Pandey, \textsuperscript{\rm 2 \#}
        Vaibhav Gautam \textsuperscript{\rm 2}
        Kanav Bhagat \textsuperscript{\rm 1}\\
        Tavpritesh Sethi \textsuperscript{\rm 1}\\
}
\affiliations {
    \textsuperscript{\rm 1} Indraprastha Institute of Information Technology Delhi \\
    \textsuperscript{\rm 2} Shiv Nadar University \\
    tavpriteshsethi@iiitd.ac.in
}

\begin{document}
\maketitle
\begin{abstract}
The COVID-19 pandemic has put immense pressure on health systems which are further strained due to the misinformation surrounding it. Under such a situation, providing the right information at the right time is crucial. There is a growing demand for the management of information spread using Artificial Intelligence. Hence, we have exploited the potential of Natural Language Processing for identifying relevant information that needs to be disseminated amongst the masses. In this work, we present a novel Cross-lingual Natural Language Processing framework to provide relevant information by matching daily news with trusted guidelines from the World Health Organization. The proposed pipeline deploys various techniques of NLP such as summarizers, word embeddings, and similarity metrics to provide users with news articles along with a corresponding healthcare guideline. A total of 36 models were evaluated and a combination of LexRank based summarizer on Word2Vec embedding with Word Mover distance metric outperformed all other models. This novel open-source approach can be used as a template for proactive dissemination of relevant healthcare information in the midst of misinformation spread associated with epidemics.
% \PACS{PACS code1 \and PACS code2 \and more}
% \subclass{MSC code1 \and MSC code2 \and more}
\end{abstract}

\section{Introduction}
Coronavirus disease (COVID-19), caused by the (SARS-Cov2) assumed pandemic proportions in March \cite{stoye2020china}. Since the outbreak of COVID-19, a massive amount of misinformation spread has taken place at an unprecedented rate through various social media platforms and other outlets \cite{vaezi2020infodemic}. This infodemic has created further difficulties in identifying the right solutions as rapidly spreading false information hampers an effective public health response. This is all the more important in lower literacy settings, where guidelines in the local language may be unavailable. This work aims to bridge this gap of providing the right information, to the right people, at the right time in the right format. \\
While several COVID-19 data resources of text were made public to users for building models, these have not been used effectively to mitigate widespread irrelevant information. This article demonstrates the feasibility of an NLP approach in mitigating the prevailing infodemic. A pipeline of various models including text summarizers, word embeddings, distance metrics, and human-in-the-loop evaluations has been developed. A total of 36 models have been evaluated in order to achieve text similarity between the news relevant to COVID-19 and WHO guidelines. The proposed pipeline follows the given workflow.
\begin{itemize}
    \item News articles and WHO guidelines related to COVID-19 are collected. 
    \item These were pre-processed to remove the unwanted characters. 
    \item These pairs were then summarized using different summarizers. 
    \item The embeddings of the summarized News articles and WHO guidelines were calculated.
    \item  The similarity between these pairs was calculated as the distance between the summarized embeddings.
\end{itemize}

Using the proposed pipeline, multiple embeddings, summarization techniques, and distance metrics are effectively used as a tool to mitigate infodemic by supplementing user's daily news consumption. This is done by providing relevant and vetted content to users from sources such as WHO along with the news. This work is first of its kind, incorporating NLP techniques in order to address the issue of widespread irrelevant information accompanying the pandemic and sets the baseline for all future works. This open-source pipeline can act as a template for addressing the misinformation fight which is inevitable in such scenarios.
\section{Methods}
In this section %we have described%
the dataset used and the techniques deployed in the pipeline have been described.

\begin{table*}[!t]
\centering
\begin{tabular}{|l|l|}
\hline
\textbf{Relevant}                                                                                                                                                                                                                                                                                                                                                                                                                                                                                                                                                                                                                                                                                                                                                                                                                                                                                 & \textbf{Irrelevant}                                                                                                                                                                                                                                                                                                                                                                                                                                                                                                                                                                                                                                                                                                                             \\ \hline
\textit{\begin{tabular}[c]{@{}l@{}}\textbf{News Article:} Are Newspapers Safe Amid Coronavirus:\\ Along with the coronavirus, rumors and myths related \\ to it are spreading fast. Which in itself is very dangerous\\ and is causing panic and fear in people. A similar myth \\ is also associated with newspapers, claiming that the last \\ can also be carriers of the coronavirus. However, according\\ to WHO information, it is safe for anyone to receive and \\ read the newspaper. It is safe to obtain any package from \\ any area where COVID-19 infections have been detected. \\ Answering questions related to the coronavirus, the WHO \\ said, An infected person is less likely to contaminate \\ commercial goods and there is less risk of infecting the \\ the virus with a packet, which rotates, is placed, and \\ exposed to various conditions and temperatures.\end{tabular}} & \textit{\begin{tabular}[c]{@{}l@{}}\textbf{News Article:} Kanika Kapoor brother Anurag has \\ spoken to his sister from London on her current \\ situation. Coronavirus infected singer is admitted\\ to a hospital in Lucknow. Kanika Kapoor \\ coronavirus test positive. The singer recently came \\ from London and went to Lucknow to be with her \\ family. Anurag told Spot Boy,  Yes, she went to \\ London and after coming back she complained of \\ a sore throat and flu. We got it tested\\ (for coronavirus) and it came back positive. Kanika\\ is currently admitted to PGI Hospital.\end{tabular}}                                                                                                                                      \\ \hline
\textit{\begin{tabular}[c]{@{}l@{}}\textbf{WHO Guideline:} It is not certain how long the virus that\\ causes COVID-19 survives on surfaces,but it seems to \\ behave like other coronaviruses. Studies suggest that \\ coronaviruses (including preliminary information on the \\ COVID-19 virus) may persist on surfaces for a few hours\\ or up to several days. This may vary under different \\ conditions (e.g. type of surface, temperature, or humidity \\ of the environment). If you think a surface may be infected,\\ clean it with simple disinfectant to kill the virus and protect\\ yourself and others. Clean your hands with an \\ alcohol-based hand rub or wash them with soap and water. \\ Avoid touching your eyes, mouth, or nose.\end{tabular}}                                                                                                                                   & \textit{\begin{tabular}[c]{@{}l@{}}\textbf{WHO Guideline:} In a situation like this it is normal \\ to feel sad, worried, confused, scared or angry. \\ Know that you are not alone and talk to someone \\ you trust, like your parent or teacher so that you\\ can help keep yourself and your school safe and\\ healthy. Ask questions, educate yourself and get \\ information from reliable sources Wash your hands \\ frequently,always with soap and water for at least 20 \\ seconds. Remember to not touch your face. Do not \\ share cups, eating utensils, food or drinks with \\ others. Model good practices such as sneezing or \\ coughing into your elbow and washing your hands, \\ especially for younger family members.\end{tabular}} \\ \hline
\end{tabular}
\caption{\label{samples} The table demonstrates a sample of relevant and irrelevant class.} 
\end{table*}

\subsection{Dataset}

The news articles dataset was scrapped from a vernacular Hindi News Portal's (Dainik Jagran) online publication. Articles presented to the annotators were in English which were translated from Hindi using Google Translate API. WHO guidelines related to COVID-19 was scrapped manually from the WHO website. WHO COVID-19 guidelines and News Articles were collected starting from February 2020 to May 2020. These news articles were filtered to remove all the non-COVID articles. The distillation was done using the following keywords `COVID’, `COVID-19’, `Corona’, `Coronavirus’, `Lockdown’, and `Pandemic’. A set of thousand random pairs were generated between news articles and the WHO guidelines. 8 annotators were assigned to annotate all 1000 pairs answering one simple question. The annotators were educated based on certain sample pairs of matching which were done among highly knowledgeable fellow researchers. Guidelines for matching the pairs were issued to the annotators for a clear perspective. The annotators were asked to determine whether the given set of pairs for news articles and WHO bulletins belong to a relevant or irrelevant class. Each annotator individually annotated each pair, thus there was no inter-user dependency and each rating was solely a single user's perspective. 

In order to clarify the scope of each class, possible cases that apply to the particular class were provided to the annotators. For instance, if the pair of WHO guidelines and news articles are contextually related, then the pair would be labeled as a relevant class. Similarly, it would be labeled as an irrelevant class if the context of the news article and WHO guidelines were not the same. For a better understanding of the guidelines and context for relevant and irrelevant classes, these classes were predefined with a set of examples. 

\begin{table}[!htp]
\centering
\begin{tabular}{lll}\hline
               & \textbf{Irrelevant} & \textbf{Relevant} \\ \hline
\textbf{Train} & 385               & 414                 \\
\textbf{Test}  & 97                & 104                 \\\hline
               & 482               & 518         \\ \hline      
\end{tabular}
\caption{\label{dataset} The dataset consists of 1,000 samples (482 Irrelevant and 518 Relevant). 
}
\end{table}

\textbf{Relevant:} A matching between a news article and WHO guideline is defined as relevant if a particular user finds the WHO guideline accompanying the news article to be related to one another. An example where all the annotators have labeled the pair of samples as relevant is given in Table \ref{samples}.

\textbf{Irrelevant:} If the user finds that the WHO guideline and the corresponding news article are unrelated, the matching is termed irrelevant. An example annotated as irrelevant by all annotators is given in Table \ref{samples}. 

The evaluation of annotations was done using the Fleiss’ Kappa metric. The Fleiss’ Kappa metric determines the level of agreement between multiple annotators and in this case, the value turned out to be 0.235, implying a fair agreement among the annotators. The scale of Kappa value with interpretation have also been provided to understand the agreement among the annotators ( $<$ 0 - No agreement, 0 — 0.20 Slight, 0.21 — 0.40 Fair, 0.41 — 0.60 Moderate, 0.61 — 0.80 Substantial, 0.81–1.0 Perfect ). The annotated pairs were later assigned binary labels 0 or 1 (irrelevant and relevant respectively) based on the majority voting for that particular pair done by the annotators. The pairs with more than 5 votes for the relevant class were labeled as 1 whereas the pairs with more than 5 votes for the irrelevant class were labeled with 0. In the case of a tie with four votes for relevant and irrelevant respectively, those samples were randomly distributed equally among both the classes so that there is no imbalance between two classes.

The new data generated after annotation contained three columns, first contained the news article, the second column contained the WHO guidelines and the third column contained the binary labels determining the relevant and irrelevant class. The data was split into test and train sets with a ratio of 80:20 for modeling purposes as shown in table \ref{dataset}. 

\subsection{Proposed Workflow}

After collection of the dataset and annotation from the annotators, pre-processing of both the news articles and the WHO guidelines was done to remove all unwanted characters. The processed data was fed into summarizers as input for generating the summaries of the news article and the WHO guidelines. The word embeddings of words in the generated summaries were calculated using Glove and Word2Vec \cite{pennington2014glove,mikolov2013distributed}. From the generated word embeddings, the weighted mean of all embedded word vectors was found using Smooth Inverse Frequency to obtain a single embedded vector representing the entire news article and WHO guidelines \cite{arora2016simple}. Distance metrics were used to find the similarity between the two generated article vectors (i.e between embedded vectors of the news article and WHO guidelines). 

The similarity score between all pairs in the training set was used to calculate the Youden's Index for each model to obtain the optimal similarity threshold, where pairs with similarity above the threshold are considered relevant and vice-versa \cite{youden1950index}. The optimal similarity threshold represents the similarity score at which the difference between the number of pairs correctly classified as relevant and the number of pairs incorrectly classified as relevant is maximized. This effectively serves the purpose of providing only relevant information to the end-user while simultaneously minimizing the delivery of irrelevant information. Further, the optimal threshold obtained for the particular model on the training set was used to evaluate the Youden's metric on the test set. The similarity scores were calculated for each model on the test pairs and pairs with similarity greater than the optimum threshold for the particular model comprised of the relevant class and pairs with similarity score below the threshold formed the irrelevant class. The original WHO guidelines and News articles were provided to the users corresponding to the relevant class. The detailed description of all the models and methodologies has been explained in the subsections below. 

\begin{figure*}[!htp]
    \centering
    \includegraphics[width=\textwidth,height=2\textheight,keepaspectratio]{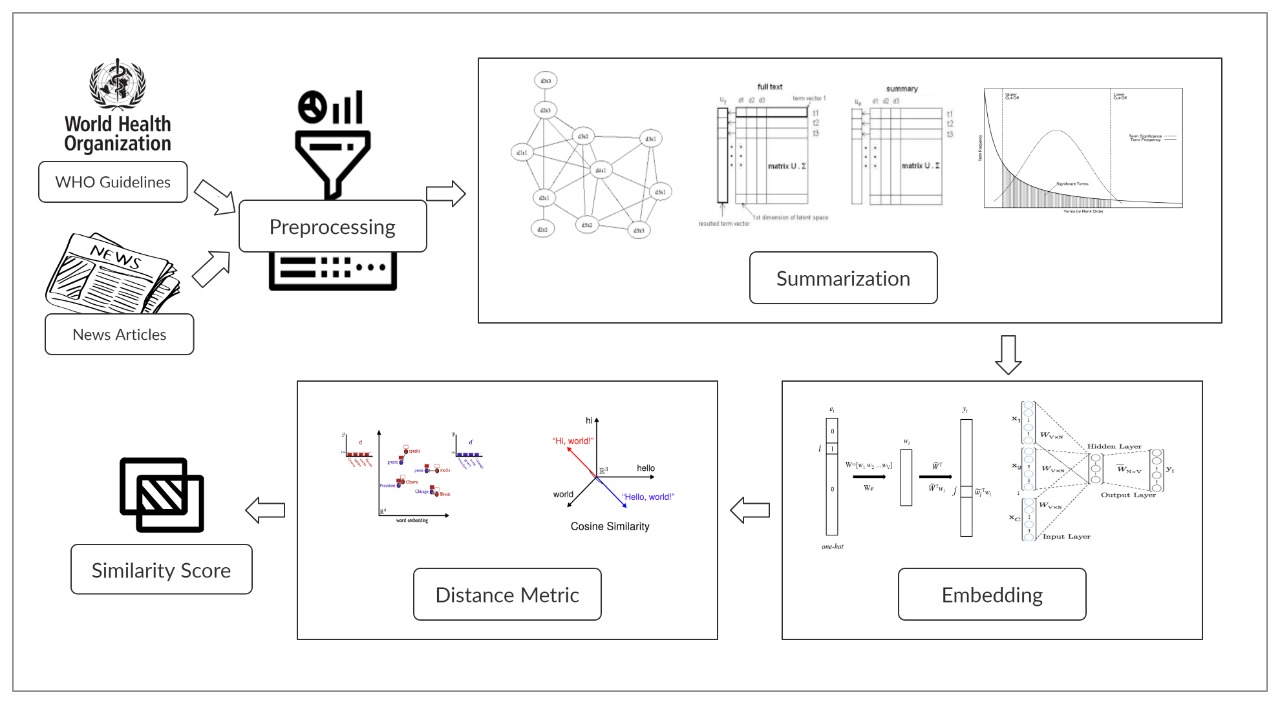}
    \caption{Proposed Framework. The pipeline takes news articles and WHO guidelines as input. These are then preprocessed, summarized, converted to embeddings and finally, the distance between the generated embeddings is delivered as the similarity score. Various combinations of summarization techniques, embedding, and distance metrics were used.}
    \label{fig:my_label}
\end{figure*}

\subsubsection{Preprocessing}
Preprocessing is a method in which text is transformed into a machine-readable format. The preprocessing pipeline in our framework includes punctuation removal, conversion of input text to lowercase, tokenization, stop word removal, and stemming algorithm. Tokenization has been done using the NLTK tokenizer. For stemming, we have used Porter stemming which seemed to be appropriate for our use case on the basis of qualitative analysis of different stemming algorithms.

\subsubsection{Summarization}

The effectiveness of numerous extractive summarization techniques has been explored in this framework to generate summaries from various lengths of news articles and WHO guidelines. This technique was used for reducing the size of the articles and guidelines while preserving the main context. These techniques prevent the introduction of errors, hence extracting meaningful sentences. This helped in removing unnecessary sections of text thereby making the framework efficient. The summarization of articles and guidelines thus enhanced the speed of the framework because the succeeding step was to generate word embeddings. The summaries of news articles and WHO guidelines contained only the important sentences and the calculation of word embeddings was done only for the words in these sentences. The following extractive text summarization techniques were used to elicit the most important sentences from these articles and guidelines. 

\textbf{\emph{Graph-Based}} - For summarization of the news articles along with WHO guidelines, the following Graph-Based summarization techniques were used: LexRank, Reduction, PyTextRank, and TextRank with Gensim \cite{erkan2004lexrank,jing2000sentence,nathan2016pytextrank,barrios2016variations}. 
All these techniques are slight variations of the primitive PageRank Algorithm. TextRank uses the typical PageRank approach while LexRank additionally uses cosine similarity of TF-IDF vectors simultaneously accounting for the position and length of sentences\cite{mihalcea2004textrank}. In these algorithms, concatenated text from the articles was split into individual sentences and embedded into vectors. The similarity score was calculated for these sentence vectors to generate a similarity matrix. The similarity matrix was then converted into a graph-based structure for sentence rank calculation, with similarity scores as the edges and sentences as the vertices of the matrix. Finally, a certain number of top-ranked sentences formed the final summary of the news articles and WHO guidelines.

\textbf{\emph{Topic-Based}} - Latent Semantic Analysis(LSA) Summarizer is a topic-based summarization technique \cite{ozsoy2011text}. This approach used a Bag-of-words to create a  term-document matrix for the news articles and WHO guidelines where rows represented the terms present in the articles and columns represented the sentences within the articles(which contained the terms). The term-document matrix is represented by X where elements \textit{(i,j)} represent the occurrence of term i in article j. 

The rows of the matrix represented vectors corresponding to a term and its relation with different sentences present in the article. The columns represented vectors corresponding to sentences and their relation with different terms present in the corpus of the individual article. A decomposition of X was done such that U and V are orthogonal matrices. This decomposition of X is called singular value decomposition(SVD). The term-document matrix was decomposed into U and V, where U and V are orthogonal matrices of X. 

\begin{equation}
    X=U\Sigma V^T
\end{equation}

Here $\sum$ is the matrix of singular values and $u_i$ and $v_i$ are left and right singular vectors. Matrix $V^T$ denotes the strength of each word in a sentence. Hence, it effectively extracted hidden semantic structures of words and sentences present in the articles. The important sentences for summaries were extracted using this Singular Value Decomposition(SVD).

\textbf{\emph{Feature-Based}} - The feature-based method of summarization used was Luhn’s method \cite{luhn1958automatic}. One of the oldest approaches, Luhn summarization began with a frequency analysis of the words in the articles. This was followed by the calculation of the weight of the sentence based on the window size of fewer frequency words between high-frequency words. The sentences with the highest weights were given as the summary for the news articles and WHO guidelines.

\textbf{\emph{Vocabulary Minimization}} - KL divergence based summarization is effectively a vocabulary minimization technique \cite{sripada2009summarization}. This method is based on minimizing the divergence of summarized vocabulary and input vocabulary, giving low KL divergence of a good summary and input document. It works by greedily adding sentences to a summary as long as it decreases the KL Divergence thus focusing on the minimization of summary vocabulary by checking the divergence from the input vocabulary. In our case, P was the original article distribution and Q was the summarized document distribution. The distribution of the summarized document was matched with the distribution of the original text, thereby minimizing the difference using the KL divergence technique. 
\begin{equation}
    KL(P,Q) = \sum_{i=1}^{n}  [ p_i * log( p_i / q_i)]
\end{equation}

\subsubsection{Embedding}

Embedding is a method in which the words from the textual data are converted into numeric vectors that can be processed by the machines for modeling. Embeddings are extremely useful in reducing the dimensionality of the textual data. The technique used for calculating article-level word embedding is stated as follows.
\begin{itemize}
    \item The word embeddings were calculated from the summary of the news articles and WHO guidelines using Word2vec or Glove. 
    \item The weighted average of these word embeddings was calculated using Smooth Inverse Frequency. 
    \item This weighted average was considered as the embedding for the summarised article.
\end{itemize}

 The following models were used for the validation of the architecture.

\textbf{\textit{Word2Vec}} - A neural network with two layers that takes text corpus as input and vectorizes it \cite{mikolov2013distributed}. It uses two methods for calculating the word representation, CBOW (Continuous Bag of Words) and Skip-Gram. The Word2Vec model was used for finding the embedding vector of words present within the article corpus. The skip gram algorithm randomly initializes vectors for these words in the corpus. The algorithm then iterates through the set of all words \textbf{T} and defines a centre word \textbf{c} at the position \textbf{t} and its context word is defined as \textbf{o}. The window size defined for context words is \textbf{m}, which means the model will consider a window of $\textbf{t+m}$ and \textbf{t-m} for searching the context words. The model then tries to maximize the likelihood \textbf{$L(\theta)$} of the context words given the center word, i.e. the probability of model predicting the context words given the center word. The likelihood function  \textbf{$L(\theta)$} is represented by : 

\begin{equation}
    L(\theta)=\prod_{t=1}^T \prod_{\substack{-m \leq j \leq m \\ j \neq 0}} P(w_{t+j} | w_t; \theta)
\end{equation}

For maximizing the likelihood function, the derivative of the function \textbf{$J(\theta)$} was taken after taking a log of the likelihood equation and multiplying it by -1 for calculating the -ve log-likelihood.

% \begin{eqnarray*}
%     J(\theta)= - \frac{1}{T}logL(\theta)\hspace{3.6cm} \\
%     = - \frac{1}{T} \sum_{t=1}^{T} \sum_{\substack{-m \leq j \leq m \\ j \neq 0}} log P(w_{t+j}| w_t;\theta)
% \end{eqnarray*}

\begin{equation}\begin{split}
               J(\theta)= - \frac{1}{T}logL(\theta)\hspace{3.6cm} \\
    = - \frac{1}{T} \sum_{t=1}^{T} \sum_{\substack{-m \leq j \leq m \\ j \neq 0}} log P(w_{t+j}| w_t;\theta)         
\end{split}\end{equation}

The words were represented by two sets of vectors, \textbf{$U_w$} and \textbf{$V_w$}. \textbf{$U_w$} is the vector of the context word and \textbf{$V_w$} is for center word. These two vectors were used for calculating the probability \textbf{P} of the center word \textbf{o} given context word \textbf{c} which is given as, 

\begin{equation}
    P(O = o | C = c ) = \frac{exp(u_o^T v_c)}{\sum_{w \epsilon Vocab} exp(u_w^T v_c)}
\end{equation} 

The loss function was minimized using the gradient descent algorithm to update the weights of the word vectors. This would lead to the vector representation of words, and similar words with the same context will point in the same direction. The key advantage of Word2Vec is that high-quality word embeddings can be learned in an efficient space and time which further allows the large embeddings with more dimensions to be learned using a larger corpus of texts. The pre-trained word vectors for the Word2Vec model were loaded from the gensim library. The version of the pre-trained embedding used for experimentation is the publicly available Word2Vec model pre-trained on Google News corpus (3 million 300-dimension English word vectors). 

\textbf{\textit{Glove}} - It is one of the word embedding models which works on the statistical theory that predicts the behavior of the words based on their co-occurrence probabilities \cite{pennington2014glove}. This helps in incorporating meaningful vectors to the words present within the vocabulary. The probability \textbf{$P_{ij}$} calculates the occurrence of word j in the context of a word i within the corpus.

\begin{equation}
    P_{ij}= P(j|i)=\frac{X_{ij}}{\sum_{k \forall context}X_{ik}}
\end{equation}

where \textbf{X} is word-word co-occurrence matrix and \textbf{$X_ij$} denotes the count of the word \textbf{j} appearing in the context of word \textbf{i}.

The function \textbf{F} built by the Glove model will predict ratios given two word vectors \textbf{$w_i$} and \textbf{$w_j$} and a context word vector \textbf{$w_k$} as inputs. 

\begin{equation}
    F(w_i,w_j, \Tilde{w_k}) = \frac{P_{ik}}{P_{jk}}
\end{equation}

The Glove model then builds an objective function that determines the relation between word vectors and text statistics. The final Glove function \textbf{J} is 

\begin{equation}
    J= \sum_{i,j=1}^{V} f(X_{ij})(w_i^T\Tilde{w_j}+b_i+\Tilde{b_j}-log(X_{ij}))^2
\end{equation}

The algorithm effectively minimizes this function to learn meaningful vector representations. The version of Glove used for experimentation is the publicly available common Crawl (840B tokens, 2.2M vocab, cased, 300 dimension vectors).

\subsection{Distance Metrics}
The similarity score between the embeddings of news articles and the WHO guidelines was found using Cosine Similarity and Word Mover Distance(WMD) \cite{lahitani2016cosine,kusner2015word}.

\begin{table*}
\small
\centering
\begin{tabular}{lllllll}
\hline
\textbf{Embedding} & \textbf{Summarizer} & \textbf{Metric} & \textbf{CutoffPoint} & \textbf{TPR} & \textbf{FPR}  & \textbf{Youden's index} \\\hline
Glove              & -                   & WMD                        & -6.267         & 0.242 & 0.135  & 0.107            \\
\textbf{W2V}                & \textbf{LexRank}             & \textbf{WMD}                        & \textbf{-2.858}         & \textbf{0.320} & \textbf{0.125}         & \textbf{0.195}            \\
Glove              & LexRank             & WMD                        & -6.427         & 0.213  & 0.083 & 0.130            \\
Glove              & PyRank              & WMD                        & -6.812         & 0.349 & 0.156       & 0.193            \\
W2V                & PyRank              & WMD                        & -2.945             & 0.281 & 0.187        & 0.094           \\
W2V                & -                   & WMD                        & -2.719         & 0.203 & 0.072 & 0.130            \\
Glove              & Luhn                & WMD                        & -6.617          & 0.485 & 0.354  & 0.131            \\
W2V                & KL                  & WMD                        & -2.907         & 0.174 & 0.083 & 0.091           \\
Glove              & KL                  & WMD                        & -6.710         & 0.320 & 0.208  & 0.112            \\
Glove              & Reduction           & WMD                        & -6.610        & 0.475 & 0.406       & 0.069          \\\hline
\end{tabular}
\caption{\label{top10results}
Best performing models on training data are evaluated on the test data. The corresponding False Positive Rate, True Positive Rate and Youden's statistic are reported. The model - Word2Vec + LexRank + WordMoverDistance reported the highest Youden's index.}
\end{table*}

\begin{figure*}[!ht]
    \centering
    \includegraphics[width=\textwidth,height=\textheight,keepaspectratio]{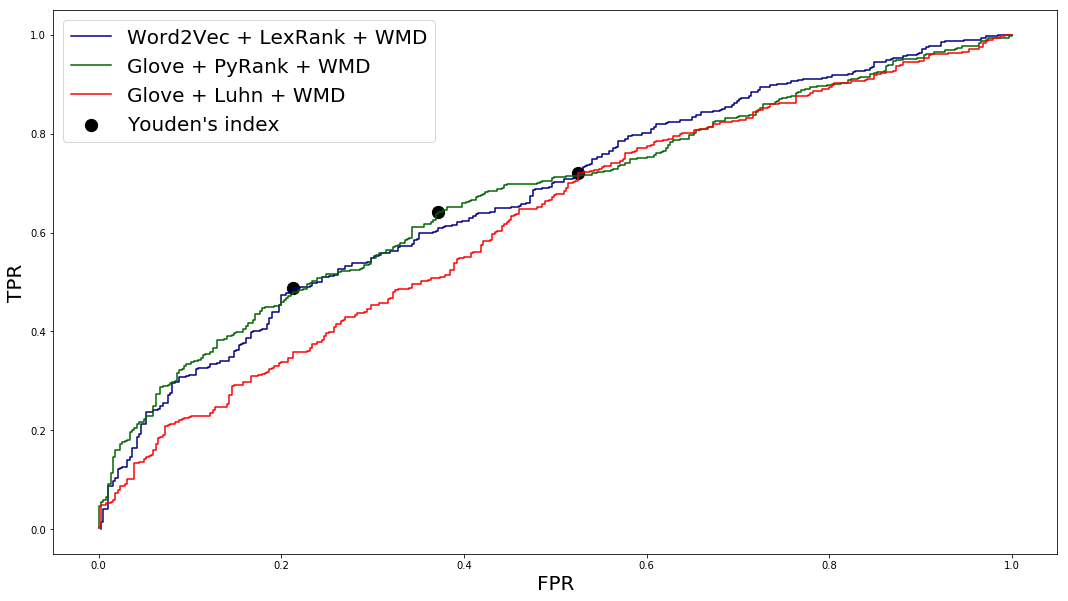}
    \caption{Sample ROC curves obtained on the training data for the the top model's. The Youden's index for each ROC curve is marked which defines the the optimum threshold for the specific model.}
    \label{fig:trainingROC}
\end{figure*}

\textbf{\emph{Cosine Similarity}} - It is used for calculating the similarity between two documents(WHO Guideline and News Article) by measuring the cosine angle formed by the two vectors. The mean weighted vectors for each guideline and news articles have been obtained using smooth inverse frequency. Cosine distance is not dependent upon the size of documents. Hence the cosine similarity of embedding vectors for News article and WHO guidelines will be close to 1 if they are similar to each other, while it will be -1 if they are completely dissimilar.

\textbf{\emph{Word Mover Distance(WMD)}} - It is used for finding the Euclidean distance of weighted embedded vectors. The Word Mover distance for the news article and WHO guidelines were calculated to find a similarity score between them. Word Mover Distance helps compare the two documents even when there is no word in common. The algorithm uses bag-of-words representation (which puts the word frequencies in the documents) finding the minimum traveling distance between articles and guidelines in an efficient way. 

\section{Results}
\subsection{Evaluation Metrics}
The objective of the proposed pipeline is two-fold - maximizing relevant content delivery and simultaneously minimizing the delivery of irrelevant content to the end-user. To facilitate this, the metrics considered throughout the experimentation are True Positive Rate, False Positive Rate, and Youden's Index \cite{youden1950index}.

True-Positive rate (TPR) also referred to as Sensitivity, measures the proportion of actual relevant pairs that are correctly classified as relevant. 

\begin{equation}
    {\displaystyle \mathrm {TPR} ={\frac {\mathrm {TP} }{\mathrm {TP} +\mathrm {FN} }}}
\end{equation}

where TP = True Positive i.e Relevant Pair of news articles and WHO guidelines correctly classified as Relevant and FN = Relevant Pair of articles incorrectly classified as Irrelevant.

False-positive ratio (FPR) measures the ratio between the number of irrelevant pairs incorrectly classified as relevant (false positives) and the total number of actual negative events. These are those pairs which should have been labeled as irrelevant, but got classified as relevant pair.

\begin{equation}
    {\displaystyle \mathrm {FPR} ={\frac {\mathrm {FP} }{\mathrm {FP} +\mathrm {TN} }}}
\end{equation}

where FP = Irrelevant Pair of news articles and WHO guidelines incorrectly classified as Relevant and TN = Irrelevant Pair of articles correctly classified as Irrelevant.

Youden's Index (J) is a metric used for analyzing ROC curves. It is defined for all points on the ROC curve and is effectively used as a criterion for the selection of optimum cut-off points in the ROC curves.

\begin{equation}
 Youden's \ Index(J)=TPR-FPR 
\end{equation}

\subsection{Experimental Results}
Multiple combinations of models have been deployed in the proposed framework. All possible combinations with a set of word embedding models (Glove and Word2Vec), summarization techniques (Summa, Gensim, Pyrank, LexRank, LSA, Luhn, KL Divergence, Reduction, and No summarizer), and distance metrics (Cosine and Word Mover Distance) have been evaluated. A combination of 36 models was generated for each possible configuration. Given the labeled dataset of Relevant and Irrelevant pairs, we deploy our models on all pairs and the evaluation is done in the following manner.

Youden’s Index based thresholding was used to find the optimum cut-off point for each of these 36 models on the training data. These models were trained on 799 news articles and WHO guidelines pair as mentioned in table \ref{dataset}. The corresponding ROC curves were plotted for each similarity score and the optimum threshold for each is obtained based on the Youden's index on the training data. The optimal threshold represents the similarity score which maximizes the difference between the TPR and FPR. It effectively implies the maximization of the number of relevant articles while minimizing the irrelevant articles delivered to the users. The values of similarity scores used to plot the ROC for each model was bound between the maximum and minimum similarity scores for the particular model.

The ROC curves obtained on the training data for the top-performing models are shown in Fig. \ref{fig:trainingROC}. Using the optimum threshold, the top 10 models with the highest Youden's index were evaluated on the test set. All pairs with a similarity score above the optimum threshold were classified as relevant and ones with scores below the threshold were classified as irrelevant. The results obtained on the test set are shown in Table \ref{top10results}. The model with the best performance on the test set used Word2Vec embedding, LexRank based summarizer, and WMD distance metric. This model had the highest value of Youden’s metric 0.1953, with a True Positive Rate of 0.320 and False Positive Rate of 0.125. Word Mover Distance outperformed Cosine Distance as the distance metric as it was common in all top 10 models. All the experimentation has been performed on Intel Core i5-7200U CPU with 8GB RAM with no GPU support. The average run time for each proposed model is ~ 300 seconds making the proposed pipeline computationally inexpensive.
\section{Discussion}
The spread of COVID-19 has been exponential where lakhs of people have been affected by the virus. Due to the rapid spread of the virus in India, there has been a fear of concern related to the disease in the economically weaker section. This has led to the spread of irrelevant information through various social platforms. 

The idea of the project was to disseminate relevant and reliable information of news articles and WHO guidelines to hard-to-reach groups staying in villages or suburban areas where the mode of communication is only Hindi. The WHO guidelines relevant to the corresponding news article will be provided to users. WHO guidelines was considered as it contains trusted and vetted forms of information related to the pandemic. A vernacular Hindi News Portal (Dainik Jagran) was chosen for scraping the news articles so that it could help in transmitting the news and relevant COVID-19 articles in regional language (Hindi) via speech translation. On qualitative evaluation, the Google Translate API used for conversion of Hindi to English retained more context in the sentence than the conversion of English to Hindi. Given that the objective was to communicate the final news articles in Hindi, an incorrect translation from English to Hindi resulted in bad user experience. In any language conversion, there is an information loss that majorly consists of stopwords and for our modeling the loss of stopwords was insignificant. Thus the news articles were converted into English which was further processed into the framework for modeling. 

Although WHO guidelines will be changing over time, the relevant and irrelevant annotations for new WHO guidelines paired with news articles will subsequently be crowdsourced from the users through an optional automated feedback form after every matching. The majority label obtained from the crowdsourced annotations will be assigned as the class label for that particular matching. The new crowdsourced dataset will now serve as the training set to designate the threshold for the new guidelines. This human-in-loop interacting framework would ensure that our model will be robust to changes in WHO guidelines and will constantly stay updated and relevant. The only shortcomings that our study had were the fact that it required manual annotation for initial training due to which the model was trained on less number of samples. Currently, the human-in-the-loop feedback chain has been introduced for annotations related to new guidelines which would be incorporated in the model subsequently for training purposes. This study will be extended for generating article-embeddings for news articles and WHO guidelines using Doc2vec \cite{le2014distributed}. Also, various other language models will be explored for creating the embeddings in order to understand the feature space article vectors.

\section{Conclusion}
The research conducted empirically tested the efficacy of NLP as an infodemic management tool. Specifically, it explored various models on extractive text summarization, word embeddings, and distance metrics. The result is an open-source AI framework for managing the spread of irrelevant information during the time of epidemics with Natural Language Processing. Incorporating WHO guidelines with relevant news articles based on user reviews, this learning model prove to be effective. Based on our knowledge from the current literature survey, there exists no such work addressing the information spectrum accompanying any healthcare debacle. From the evaluation, it was found that LexRank along with Word2vec and Word Distance Mover proved to be the best model for matching WHO guidelines with news articles related to COVID-19 in a relevant manner. Being the first of it's kind, the proposed architecture was evaluated based on the manually annotated data using Youden’s index. This framework can effectively be incorporated into existing channels of information propagation. Given the limited data available for training, the promising results highlight the tremendous scope for such models in addressing the glut of irrelevant information spread.

\section*{Declarations}
\subsection{Funding}
The authors did not receive support from any organization for the submitted work.

\subsection{Conflict of interest}
The authors declare that they have no conflict of interest.

\subsection{Availability of data and material}
The datasets generated during and/or analysed during the current study are available from the corresponding author on reasonable request.

\subsection{Code availability}
The code is available from the corresponding author on reasonable request.
\bibliography{aaai21.bib}

\end{document}